\documentclass{article}
\usepackage{arxiv}
\usepackage[utf8]{inputenc} %
\usepackage[T1]{fontenc}    %
\usepackage{hyperref}       %
\usepackage{url}            %
\usepackage{booktabs}       %
\usepackage{amsmath,amssymb,amsfonts}       %
\usepackage{nicefrac}       %
\usepackage{microtype}      %
\usepackage{lipsum}		%
\usepackage{graphicx}

\usepackage[english]{babel}
\usepackage[caption=false,font=footnotesize]{subfig}
\usepackage[center]{caption}
\usepackage{textcomp}
\usepackage[table]{xcolor}
\usepackage{hhline}
\usepackage{multirow}
\usepackage{pgf}
\usepackage{collcell}

\def\colorModel{hsb} %
\newcommand\ColCell[1]{
  \pgfmathparse{#1<50?1:0}  %
    \ifnum\pgfmathresult=0\relax\color{white}\fi
  \pgfmathsetmacro\compA{0}      %
  \pgfmathsetmacro\compB{#1/100} %
  \pgfmathsetmacro\compC{1}      %
  \edef\x{\noexpand\centering\noexpand\cellcolor[\colorModel]{\compA,\compB,\compC}}\x #1
  } 
\newcolumntype{E}{>{\collectcell\ColCell}m{0.325cm}<{\endcollectcell}}  %

\renewcommand{\arraystretch}{1.25}

\providecommand{\keywords}[1]
{
  \small	
  \textbf{\textit{Keywords---}} #1
}

\begin{document}
\title{Cost-Sensitive Regularization for Diabetic Retinopathy Grading from Eye Fundus Images}

\author{Adrian Galdran$^{1,*}$, José Dolz$^{2}$, Hadi Chakor$^{3}$, Hervé Lombaert$^{2}$, Ismail Ben Ayed$^{2,4}$  \\[3mm]
        $^{1}$University of Bournemouth, UK  \hspace{0.75cm} $^{3}$École de Technolgie Superieure de Montréal, Canada\\[1mm] 
      \hspace{0.75cm}   $^{4}$Diagnos Inc., Canada   \hspace{1.5cm} $^{5}$CRCHUM University of Montreal Hospital Centre \\[1mm]
        $^{*}$ Corresponding author: agaldran@bournemouth.ac.uk
}

\maketitle
\begin{abstract}
Assessing the degree of disease severity in biomedical images is a task similar to standard classification but constrained by an underlying structure in the label space.
Such a structure reflects the monotonic relationship between different disease grades. 
In this paper, we propose a straightforward approach to enforce this constraint for the task of predicting Diabetic Retinopathy (DR) severity from eye fundus images based on the well-known notion of Cost-Sensitive classification. 
We expand standard classification losses with an extra term that acts as a regularizer, imposing greater penalties on predicted grades when they are farther away from the true grade associated to a particular image.
Furthermore, we show how to adapt our method to the modelling of label noise in each of the sub-problems associated to DR grading, an approach we refer to as Atomic Sub-Task modeling. This yields models that can implicitly take into account the inherent noise present in DR grade annotations. Our experimental analysis on several public datasets reveals that, when a standard Convolutional Neural Network is trained using this simple strategy, improvements of 3-5\% of quadratic-weighted kappa scores can be achieved at a negligible computational cost.
Code to reproduce our results is released at \url{github.com/agaldran/cost_sensitive_loss_classification} .
\end{abstract}
\vspace{0.15cm}
\keywords{Diabetic Retinopathy Grading \and Cost-Sensitive Classifiers \and Label Noise}
\vspace{0.5cm}

\section{Introduction}
Diabetes is regarded as a global eye health issue, with a steadily increasing world-wide affected population, expected to reach 630 million individuals by 2045 \cite{noauthor_diabetes_nodate}. Diabetic Retinopathy (DR) is a complication of standard diabetes,
caused by damage to vasculature within the retina. DR shows early signs in the form of swelling micro-lesions that destroy small vessels and release blood into the retina. Advanced DR stages are characterized by the appearance of more noticeable symptoms, \textit{e.g.} proliferation of neo-vessels, leading to the detachment of the retinal layer and eventually permanent sight loss. 

\begin{figure*}[t]
\centering
\subfloat[]{\includegraphics[width = 0.24\textwidth]{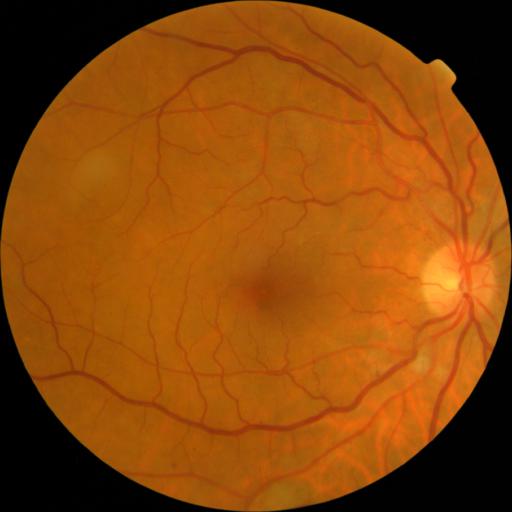}
\label{fig_deg_1}}
\hfil
\subfloat[]{\includegraphics[width = 0.24\textwidth]{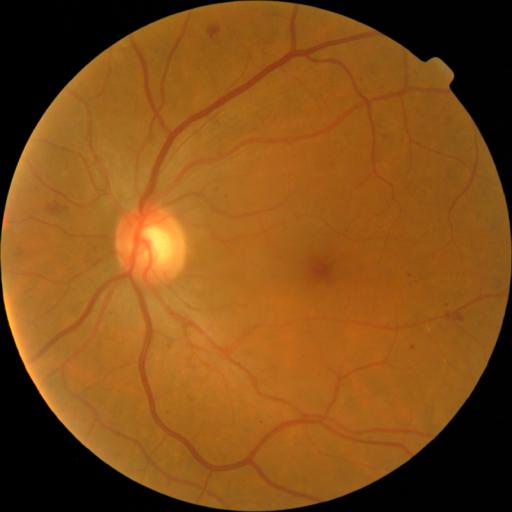}
\label{fig_deg_2}}
\hfil
\subfloat[]{\includegraphics[width = 0.24\textwidth]{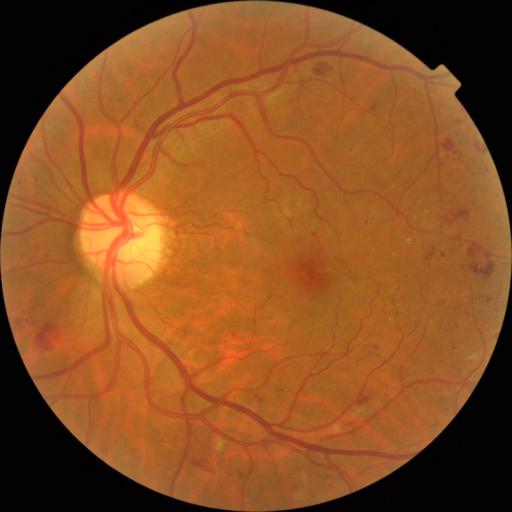}
\label{fig_deg_1}}
\hfil
\subfloat[]{\includegraphics[width = 0.24\textwidth]{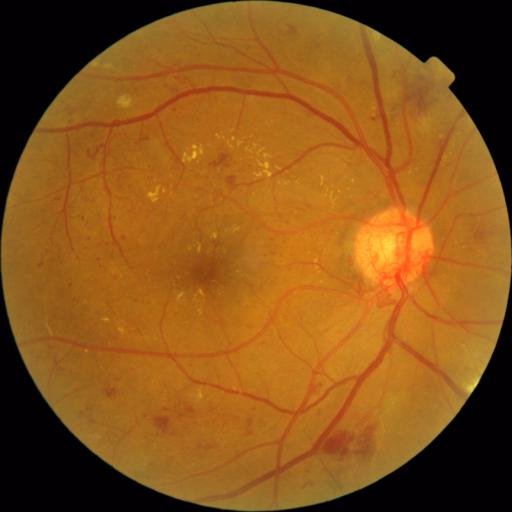}
\label{fig_deg_2}}
\caption{Images from the Messidor-2 dataset illustrating the progressive behavior of DR. (a) Grade 1 (Mild NPDR): only few microaneurysms can be found (b) Grade 2 (Moderate NPDR): Presence of multiple microaneurysms, blot hemorrhages, venous beading, and/or cotton wool spots (c) Grade 3 (Severe NPDR): Micro-aneurysms if 4 quadrants of the retina, cotton wool spots, venous beading, severe intra-retinal microvascular abnormalities. (d) Grade 4 (PDR): Neovascularization, vitreous hemorrhages.}
\label{fig_dr}
\end{figure*}

Retinal images acquired with fundus cameras are the tool of choice for discovering these early symptoms, representing an effective diagnostic tool suitable for automatic diagnostic systems \cite{valentina_bellemo_artificial_2019}. 
In this context, and with the advent of Deep Learning in the last decade, a wide set of techniques has been proposed in recent years \cite{gulshan_development_2016,abramoff_improved_2016,costa_weakly-supervised_2018}. 
However, the vast majority of these works are designed for the screening task, \textit{i.e.} distinguishing healthy individuals from patients at any stage of risk. 
Due to its difficulty, fewer works have addressed the task of DR grading, consisting of classifying an eye fundus image into one of the five categories proposed by the American Academy of Ophthalmology \cite{wilkinson_proposed_2003}, illustrated in Fig. \ref{fig_dr}. 
In addition, most recent DR grading techniques \cite{krause_grader_2018,li_automatic_2019,sahlsten_deep_2019} have focused on scaling up existing Convolutional Neural Networks by considering larger/better databases, but only a few works addressed the design of customized loss functions that are more suitable for this task, which is the goal of this paper.

Cost-Sensitive classifiers are known to be useful for addressing two of the main challenges related to DR grading. First, they allow to model the underlying structure of an heterogeneous label space \cite{frogner_learning_2015,mensch_geometric_2019,lin_cost-sensitive_2019}. Second, they are beneficial for dealing with severely class-imbalanced scenarios \cite{thai-nghe_cost-sensitive_2010,zhi-hua_zhou_training_2006}.
Despite this, to the best of our knowledge, no previous work has explored Cost-Sensitive loss minimization approaches in the context of DR grading from eye fundus images. 

In this paper, we present a straightforward approach for integrating Cost-Sensitive classification constraints in the task of DR grading from retinal images. 
We choose to introduce these constraints by attaching an auxiliary Cost-Sensitive loss term to popular miss-classification error functions, and by analyzing the impact of this process in the training of a standard CNN.
In addition, we illustrate how to adapt our method to the modeling of label noise in each of the sub-problems associated to DR grading, an approach we refer to as {\em Atomic Sub-Task modeling}.
We conduct a series of careful experiments demonstrating that expanding well-known loss functions with a Cost-Sensitive term brings noticeable performance increases, and that sub-task modeling leads to learning models that behave more similarly to human annotators.

\section{Methodology}
In this section we first describe our approach to build Cost-Sensitive (CS) classifiers, and the loss functions we select as baselines, to which we will add a CS-regularizing term. 
We then show how CS can be employed to model label noise for DR grading problems, and detail the training process we followed to optimize the parameters of our models.

\subsection{Cost-Sensitive Regularization}
In order to induce a different penalty for each kind of error, let us first consider the case in which a model $U$ produces a prediction $U(x)=\hat{y} \in [0,1]\times \displaystyle \ldots \times[0,1]$. Such prediction is to be compared with the corresponding label $y$. 
For the sake of readability, in the following we will abuse notation and refer by $y$ indistinctly to an integer label $y\in\mathbb{L}=\{1,2,3,4,5\}$ and its one-hot-encoded counterpart $y\in \{0,1\}\times \displaystyle \ldots \times\{0,1\}$, which takes a value of $1$ in the position corresponding to $y$ and $0$ elsewhere.

Standard loss functions like the cross-entropy error, described by:
\begin{equation}
\mathcal{L}_{CE}(\hat{y},y)= -\sum_{i=1}^{n=5} y_i \log(\hat{y}_i)
\end{equation}
are insensitive to any underlying structure in the label space $\mathbb{L}$. This means that for a particular example $(x,y_j)$, if any permutation is applied on $\mathbb{L}\setminus\{y_j\}$, the resulting error will remain the same. 
In order to modify that behavior, we consider a cost matrix $M$ that encodes a null cost for a prediction such that $\hat{y}=y_j$, but cost that increases along with the distance $\|y - \hat{y}\|$. 

A simple approach to achieve such increasing label-dependent penalty is by encoding in each row of $M$ those costs, and then computing the scalar product of $\hat{y}$ with the row of $M$ corresponding to $y$, i.e. $\mathcal{L}(y,\hat{y})=\langle M(y,\cdot), \hat{y} \rangle$. 
However, due to the high imbalance of the DR grading problem (with typically few examples of classes DR1, DR3, and DR4) in our experiments we noted that simply minimizing such quantity would lead to models remaining stuck in local minima and classifying all images into DR0 and DR2 classes. For this reason, we prefer to combine a CS term with a base loss as follows:
\begin{equation}\label{cs_loss}
\mathcal{L}^{cs}(\hat{y},y) = \mathcal{L}^{base}(\hat{y},y) + \lambda \langle M^{(2)}(y,\cdot), \hat{y} \rangle, \ \ \  M^{(2)}_{ij}=\|i-j\|_2^2.
\end{equation}
In the above equation, we have selected the $L^2$-based ground cost matrix $M^{(2)}$, since it fits nicely with the goal of maximizing quadratic-weighted kappa score, but other cost matrices could be easily implemented if previous knowledge of the problem is available to be embedded in the loss function. 
We give an example of how to build different penalties in the next section.

As for the base loss, in this paper we consider three different alternatives, namely the above Cross-Entropy loss together with the Focal Loss and Non-Uniform Label Smoothing Loss functions. 
The Focal Loss was introduced for object detection tasks in \cite{lin_focal_2020}, but it has become widely popular in classification tasks due to its ability to penalize wrongly miss-classified examples during training. 
In a multi-class setting, it is given by the following equation:
\begin{equation}
\mathcal{L}_{FL}(\hat{y},y)=-\sum_{i=1}^{n=5} y_i \alpha (1-\hat{y}_i)^\gamma \cdot \log(\hat{y}_i),
\end{equation}
being $\alpha$ a weighing factor and $\gamma$ the so-called focusing parameter that penalizes errors in wrongly classified examples more than errors in correctly classified ones.

Non-Uniform Label Smoothing Loss is a straightforward modification of the popular Label Smoothing technique in which neighboring labels receive more probability mass than farther-away ones. 
This process is described by the following formula \cite{galdran_non-uniform_2020}:
\begin{equation}
\mathcal{L}_{NULS}(\hat{y},y)= \mathcal{L}_{CE}(\hat{y},G_\sigma({y})),
\end{equation}
where actual labels are manipulated by means of convolution with a Gaussian kernel $ G_\sigma$ resulting in the introduction of lower penalty in neighboring grades and greater loss value for far away predictions. 
Differently from the Cross-Entropy and the Focal loss, the Non-Uniform Label Smoothing strategy is sensitive to the label space structure. 
Yet, we hypothesize that further imposing greater penalty on farther away grades could bring benefits training based on this loss, as well as the other two above functions. 
In our experiments, described below, we train several models by considering $\mathcal{L}^{base}$ to be $\mathcal{L}_{CE}$, $\mathcal{L}_{FL}$, and $\mathcal{L}_{NULS}$ and varying the $\lambda$ hyper-parameter from $\lambda=0$ (no CS regularization whatsoever) to greater CS penalty, and observe the resulting performance.

\subsection{Atomic Sub-Task Modeling}
Annotating retinal images regarding the level of DR severity is know to be a noisy process, with high rates of inter-observer disagreement \cite{krause_grader_2018,voets_reproduction_2019}. 
In this paper we propose to leverage available data regarding the structure of that disagreement to improve DR grading accuracy. 
Our hypothesis is that if the kind of noise affecting labels in the training data can be estimated, we can make a model aware of such noise via a CS mechanism similar to the one described in eq. (\ref{cs_loss}). 

Specifically, we consider the confusion matrix $M_{opht}$ from the left hand side of eq. (\ref{conf_mat_eqq}). 
This matrix contains information collected in \cite{krause_grader_2018} regarding inter-observer disagreement between retinal specialists and an adjudicated consensus during the grading process of their clinical validation dataset. 
Interestingly, this matrix conveys not only information about which grades are most likely to be subject of expert disagreement, but it also tells us which grades are more often mistaken by which other grades. 

To formalize the above, we refer to the task of categorizing an image of actual DR grade $i$ image into the $j$-th grade as $t_{ij}$, and we refer to this process as \textit{atomic sub-tasks}. 
For a given grade $D_i$, the amount of images actually belonging to that grade is $s_i=\sum_{j=1}^{n=5} t_{ij}$, and normalizing $t_{ij}$ by $s_i$ provides an estimate of $t_{ij}=P(D_j\vert D_i)$, which denotes the likelihood that an annotator diagnoses an image as grade $D_j$ when it actually was of grade $D_i$, as shown in the right hand side of eq. (\ref{conf_mat_eqq}):
\begin{equation}\label{conf_mat_eqq}
\displaystyle
M_{opht} = \begin{bmatrix}
1469 &  4 &   5 &  0 &  0\\
58   & 62 &   5 &  0 &  0\\
22   &  3 & 118 &  1 &  0\\
 0   &  0 &  13 & 36 &  1\\
 0   &  0 &   0 &  1 & 15\\
\end{bmatrix}
\ \ M_{opht}^* = \begin{bmatrix}\displaystyle
0.994 &  0.003 &   0.003 &      0 &  0\\
0.464 &  0.496 &   0.040 &      0 &  0\\
0.153 &  0.021 &   0.819 &  0.007 &  0\\
 0    &  0     &   0.260 &  0.720 &  0.020\\
 0    &  0     &       0 &  0.06237 &  0.937\\
\end{bmatrix}
\end{equation}
We assume below that matrices are indexed starting from $0$, \textit{i.e.} $0\leq i,j\leq 4$. By observing $M_{opht}^*$ we can draw several conclusions, for example:
\begin{itemize}
\item Annotators are likely to be greatly accurate when grading $D_0$ and $D_4$ images, as derived from $t_{0,0}\approx 1$ and $t_{4,4}\approx 0.94$.
\item Around $50\%$ of $D_1$ images are likely to be incorrectly labeled ($t_{11}\approx0.5$). 
\item Only $8\%$ of incorrectly labeled $D_1$ images are likely to be labeled as $D_2$.
\item Approximately $93\%$ of those incorrectly labeled $D_1$ images are likely to be labeled as $D_0$.
\end{itemize} 
Under the hypothesis that in a dataset labeled by a single annotator the reliability of the annotations will follow a distribution similar to the above, we can assume, for instance, that such dataset will contain reliable labels concerning $D_0$ grades. 
However, we may also assume that when an image has been annotated as of grade $D_1$, this is quite likely to be incorrect, and it may well be the case that such image is actually of grade $D_0$, since the corresponding atomic sub-task $t_{10}=P(D_0\vert D_1)$ holds value comparable to $t_{11}=P(D_1\vert D_1)$.

Our goal is to impose in our models a penalty on erroneous predictions that takes into account all the above information. 
That is, we want to penalize incorrect predictions when the label is likely to be reliable, but we are willing to be more \textit{tolerant} with erroneous predictions if we know the associated label is unreliable. 
Embedding this knowledge into a loss function is easily accomplished using the CS loss formulation as developed in the previous section: we consider $I-M_{opht}^*$ in eq. (\ref{cs_loss}), being $I$ the identity matrix.
Higher values of $t_{ij}$ will result in lower penalties, whereas lower values lead to a greater penalty.
 
Note, however, that for grades such that $t_{ij} = 0$, $M_{opht}^*$, there is no useful information in terms of relative reliability of these grades, e.g. $t_{03}=t_{04}=0$ does not convey the information that it is harder to misdiagnose a $D_0$ images as $D_3$ than it is to misdiagnose it as $D_4$.  
In those situations it might be better to rely on the penalty imposed by $M^2_{ij}$ from eq. (\ref{cs_loss}). 
For this reason, we suggest to implement an averaged Cost-Sensitive regularizer as:
\begin{equation}\label{astmod}
\mathcal{L}^{cs}(\hat{y},y) = \mathcal{L}^{base}(\hat{y},y) + \lambda\langle \hat{y}, M(y,\cdot)\rangle, \ \ \  M=(M^{(2)}+I-M_{opht}^*)/2
\end{equation}
We now describe the remaining training specifications aside of the loss functions.

\subsection{Training Details}
For analyzing the impact of minimizing CS-regularized loss functions in the problem of DR grading, we follow the process of varying the $\lambda$ hyper-parameter in eq. (\ref{cs_loss}). 
For each base loss function, we train a Convolutional Neural Network (CNN) by setting $\lambda=0$ (no regularization), $\lambda=0.1$, and $\lambda=1$. 
If the best performance of these three experiments results from employing $\lambda=1$, we set $\lambda=10$ and train the CNN again. 
This process is repeated until performance does not improve anymore.

As for the CNN, we select the Resnext50 architecture based on its excellent classification accuracy in other multi-class problems \cite{xie_aggregated_2017}, and weights are initialized from training in the ImageNet dataset.
We use Stochastic Gradient Descent with a batch size of $8$, and the learning rate is set to $0.001$. 
Performance (quadratic kappa score) is monitored in an independent validation set. 
The learning rate is decreased by a factor of $10$ whenever performance stagnates in the validation set, and the training is stopped after $10$ epochs of no further improvement. 
In addition, to mitigate the impact of class imbalance, we oversample minority classes \cite{buda_systematic_2018}.

\section{Experimental Validation}
In this section we describe the experimental setting we follow in order to validate our approach: considered datasets, comparing techniques, and numerical results.
\subsection{Experimental Details}
We consider as our primary dataset the Eyepacs database\footnote{\url{https://www.kaggle.com/c/diabetic-retinopathy-detection}} the largest public dataset with DR grading labels for DR grading labels. 
It contains around 80,000 high resolution retinal fundus images (approximately 35,000 are assigned to the training set, from which we employ $10\%$ for validation, and 55,000 are held out for testing purposes). 
The Eyepacs dataset contains a considerable amount of low quality images and label noise \cite{voets_reproduction_2019}. 
Therefore, it represents an interesting test-bed to observe the robustness of DR grading algorithms.

As a secondary test set, we also consider the Messidor-2 dataset \cite{abramoff_improved_2016}, which contains $1748$ images corresponding to $874$ patients. 
In this case, we employ the ground-truth labels released by \cite{krause_grader_2018}, available online\footnote{\url{https://www.kaggle.com/google-brain/messidor2-dr-grades}}.
These labels are extracted from a process of consensus adjudication of three retinal specialists, and they are therefore of much better quality than the Eyepacs dataset ground-truth.

For performance assessment, we apply as the main metric of interest the quadratic-weighted kappa score (quad-kappa), which is typically used to assess inter-observer variability, and is very popular metric in this task.  
As further measures of correlation, we also analyze Average of Classification Accuracy (ACA, the mean of the diagonal in a normalized confusion matrix \cite{zhao_bira-net_2019}) and the Kendall-$\tau$ coefficient.
We also report the mean Area Under the Receiver-Operator Curve in its multi-class extension, after considering each possible class pair\cite{hand_simple_2001}. 
For statistical testing, expert labels and model predictions in each of both test sets (Eyepacs and Messidor-2) are bootstrapped \cite{bertail_bootstrapping_2009} (n=1000) in a stratified manner with respect to the relative presence of DR grades. 
Performance differences for each metrics are derived in each bootstrap and p-values are computed for testing significance. 
The statistical significance level was set to $\alpha = 0.05$ in each case.

For comparison purposes, we select three other recent techniques that introduce methods specifically developed to solve the DR grading task: $DR\vert graduate$ \cite{araujo_drgraduate_2020}, Bilinear Attention Net for DR Grading (Bira-Net) \cite{zhao_bira-net_2019}, and Quadratic-Weighted Kappa Loss (QWKL) \cite{de_la_torre_weighted_2018}.

\subsection{Numerical Results}
After training a CNN by minimizing each of the three considered base losses (Cross-Entropy, Focal Loss, and Non-Uniform Label Smoothing) with different degrees of regularization, we select the best model and compute results first on the Eyepacs test set. 
We denote the unregularized models by CE, FL, and NULS respectively, and their regularized counterparts as CE-CS, FL-CS, and NULS-CS. 

We then select the best hyperparameter setting for each regularized model ($\lambda=10$ in all cases), and retrain the same model but this time using our proposed Atomic Sub-Task modeling, denoted by an $AST$ suffix in each case. 
We compile in Table \ref{tab_results1} the obtained results in terms of quadratic $\kappa$-score, mean AUC, ACA and Kendall-$\tau$, for all the described options.

\begin{table}[t]  %
	\renewcommand{\arraystretch}{1.3}	
	\centering
\setlength\tabcolsep{8.5pt}	
\begin{tabular}{l c c c c}
  & \textbf{quad-kappa} & \textbf{mAUC} & \textbf{ACA} & \textbf{Kendall-$\tau$} \\
\cmidrule(lr{.0005em}){1-5} 
\textbf{CE}             & 75.76$\pm$ 0.31          & 87.35$\pm$ 0.14           &  51.32$\pm$ 0.44          &  67.35$\pm$ 0.31\\
\textbf{CE-CS}          & 77.27$\pm$ 0.30          & 88.42$\pm$ 0.14           &  53.26$\pm$ 0.42          &  \textbf{69.89$\pm$ 0.30}  \\
\textbf{CE-AST}         & \textbf{77.39$\pm$ 0.29} & 88.49$\pm$ 0.13           &  \textbf{54.12$\pm$ 0.44} &  69.25$\pm$ 0.30  \\
\cmidrule(lr{.0005em}){1-5} 
\textbf{Focal Loss}     & 74.72$\pm$ 0.34          & 86.63$\pm$ 0.16           &  51.90$\pm$ 0.44          &  65.38$\pm$ 0.32\\
\textbf{FL-CS}          & 77.38$\pm$ 0.31          & 88.58$\pm$ 0.14           &  54.11$\pm$ 0.45          &  69.45$\pm$ 0.30 \\
\textbf{FL-AST}         & \textbf{77.94$\pm$ 0.29} & \textbf{88.90$\pm$ 0.13}  &  \textbf{54.71$\pm$ 0.43} &  \textbf{70.45$\pm$ 0.29} \\
\cmidrule(lr{.0005em}){1-5} 
\textbf{NULS}           & 77.09$\pm$ 0.30          & 88.44$\pm$ 0.14           &  53.02$\pm$ 0.46          &  69.47$\pm$ 0.29 \\
\textbf{NULS-CS}        & 77.91$\pm$ 0.30          & 88.82$\pm$ 0.14           &  54.55$\pm$ 0.44          &  70.14$\pm$ 0.30 \\
\textbf{NULS-AST}       & \textbf{78.71$\pm$ 0.28} & \textbf{89.05$\pm$ 0.13}  &  54.57$\pm$ 0.46          &  \textbf{71.04$\pm$ 0.30} \\[0.25cm]
\end{tabular}
\caption{Performance comparison when training without regularization, with CS regularization as in eq. (\ref{cs_loss}), and with Atomic Sub-Task modeling (AST) as in eq. (\ref{astmod}), for the three considered loss functions. Statistically significant results  are marked bold.}
\label{tab_results1}
\end{table}

Finally, we report in table \ref{tab_results3} the performance of our best model (using as a base loss NULS and Atomic Sub-Task modeling) in comparison with the techniques proposed in \cite{araujo_drgraduate_2020}, \cite{de_la_torre_weighted_2018}, and \cite{zhao_bira-net_2019}, in the test set of both Eyepacs and Messidor. 
We also provide confusion matrices for the Eyepacs test set in Fig. \ref{fig_conf_mats}.

\begin{table}[h]  %
	\renewcommand{\arraystretch}{1.3}	
	\centering
\setlength\tabcolsep{1pt}	
\begin{tabular}{l c c c c }
                               & $DR\vert graduate$ \cite{araujo_drgraduate_2020}& \small{QWKL \cite{de_la_torre_weighted_2018}} & Bira-Net \cite{zhao_bira-net_2019}  & NULS-AST \\
\cmidrule(lr{.0005em}){1-5} 
\textbf{Eyepacs}               & 74.00/53.6  & 74.00/n.a.  & n.a./54.31  & \textbf{78.71$\pm$0.28}/\textbf{54.57$\pm$0.46} \\
\textbf{Messidor-2}            & 71.00/59.60 & n.a./n.a.   & n.a./n.a.  & \textbf{79.79$\pm$1.03}/ \textbf{63.41$\pm$1.99}  \\[0.25cm]
\end{tabular}
\caption{Performance comparison in terms of quad-kappa/ACA for different methods when tested on the Eyepacs and Messidor-2 datasets. Models were trained on Eyepacs and tested on Eyepacs and Messidor (without retraining/fine-tuning).}
\label{tab_results3}
\end{table}

\newcommand\items{5}   %
\arrayrulecolor{white} %
\begin{figure*}[h]
\centering
\subfloat[]{
\begin{tabular}{cc*{\items}{|E}|}
\multicolumn{1}{c}{} & 
\multicolumn{1}{c}{\rotatebox{0}{T/P}} & 
\multicolumn{1}{c}{\rotatebox{90}{DR0}} & 
\multicolumn{1}{c}{\rotatebox{90}{DR1}} & 
\multicolumn{1}{c}{\rotatebox{90}{DR2}} & 
\multicolumn{1}{c}{\rotatebox{90}{DR3}} & 
\multicolumn{1}{c}{\rotatebox{90}{DR4}} 
\\ \hhline{~*\items{|-}|}
&DR0   & 81 & 16  & 3  & 0  & 0    \\ \hhline{~*\items{|-}|}
&DR1   & 35 & 48 & 17 & 0  & 0    \\ \hhline{~*\items{|-}|}
&DR2   & 11  & 25 & 49 & 14 & 1    \\ \hhline{~*\items{|-}|}
&DR3   & 2  & 5  & 34 & 55 & 4    \\ \hhline{~*\items{|-}|}
&DR4   & 2  & 8  & 22 & 33 & 35    \\ \hhline{~*\items{|-}|}
\end{tabular}
}
\hfill
\subfloat[]{
\begin{tabular}{cc*{\items}{|E}|}
\multicolumn{1}{c}{} & 
\multicolumn{1}{c}{\rotatebox{0}{T/P}} & 
\multicolumn{1}{c}{\rotatebox{90}{DR0}} & 
\multicolumn{1}{c}{\rotatebox{90}{DR1}} & 
\multicolumn{1}{c}{\rotatebox{90}{DR2}} & 
\multicolumn{1}{c}{\rotatebox{90}{DR3}} & 
\multicolumn{1}{c}{\rotatebox{90}{DR4}} 
\\ \hhline{~*\items{|-}|}
&DR0   & 79 & 6  & 15  & 0 & 0    \\ \hhline{~*\items{|-}|}
&DR1   & 51 & 19 & 30 & 0  & 0    \\ \hhline{~*\items{|-}|}
&DR2   & 16  & 4 & 69 & 9 & 2    \\ \hhline{~*\items{|-}|}
&DR3   & 3  & 1  & 43 & 47 & 6    \\ \hhline{~*\items{|-}|}
&DR4   & 0  & 1  & 22 & 19 & 58    \\ \hhline{~*\items{|-}|}
\end{tabular}
}                    
\hfill
\subfloat[]{
\begin{tabular}{cc*{\items}{|E}|}
\multicolumn{1}{c}{} & 
\multicolumn{1}{c}{\rotatebox{0}{T/P}} & 
\multicolumn{1}{c}{\rotatebox{90}{DR0}} & 
\multicolumn{1}{c}{\rotatebox{90}{DR1}} & 
\multicolumn{1}{c}{\rotatebox{90}{DR2}} & 
\multicolumn{1}{c}{\rotatebox{90}{DR3}} & 
\multicolumn{1}{c}{\rotatebox{90}{DR4}} 
\\ \hhline{~*\items{|-}|}
&DR0   & 97 & 2  & 1  & 0 & 0    \\ \hhline{~*\items{|-}|}
&DR1   & 68 & 24 & 8 & 0 & 0    \\ \hhline{~*\items{|-}|}
&DR2   & 26 & 13  & 50 & 10 & 1    \\ \hhline{~*\items{|-}|}
&DR3   & 4 & 2  & 39 & 52 & 3    \\ \hhline{~*\items{|-}|}
&DR4   & 7 & 1  & 18 & 24 & 50    \\ \hhline{~*\items{|-}|}
\end{tabular}
}
\caption{(a)--(c): Normalized confusion matrices corresponding to: (a) the method of Araújo \textit{et al.} \cite{araujo_drgraduate_2020}, (b) Zhao \textit{et al.} \cite{zhao_bira-net_2019}, (c) NULS-AST.}
\label{fig_conf_mats}
\end{figure*}

\section{Discussion and Conclusion}
Results on Table \ref{tab_results1} clearly show that introducing Cost-Sensitive regularization results in noticeable improvements, particularly when measuring performance in terms of quadratic $\kappa$-score. 
This is meaningful since the considered cost matrix was selected so as to quadratically penalize distance in the label space for erroneous predictions.
Quadratic $\kappa$-score experimented an improvement ranging from $3.5\%$ when regularizing the Focal loss to $1\%$ for NULS. 
This could also be expected, since NULS already introduces some asymmetry in the way DR grades are treated. 
If Atomic Sub-Task modeling is considered, these improvements are even greater when compared with unregularized counterparts: from an increase of $\kappa$ score of $4.3\%$ for the Focal Loss to an increase of $2\%$ for NULS.
It is also worth noticing that the confusion matrix resulting from training with Atomic Sub-Task modeling shows certain similarity with respect to the inter-observer disagreement matrix in the left-hand side of eq. (\ref{conf_mat_eqq}), specially when compared with the confusion matrices produced by other techniques, as shown in Fig. (\ref{fig_conf_mats}).

It should be stressed that performance on Table \ref{tab_results1} is not comparable to results of the competition that published the data. 
There are several reasons for this: the heuristics for ranking optimization common to these competitions, or the fact that participants were allowed to submit predictions on $20\%$ of the testing data during the competition.
In addition, the lack of cross-dataset experimentation complicates evaluating generalization ability. 
In contrast, the approach proposed here is a general improvement over standard techniques, not limited to the DR grading problem, and which generalizes to other datasets, as Table \ref{tab_results3} shows.

\bibliographystyle{unsrt} 
\bibliography{miccai_cs_dr}

\end{document}